\documentclass[10pt,twocolumn,letterpaper]{article}

\usepackage{cvpr}

\usepackage{times}
\usepackage{epsfig}
\usepackage{graphicx}
\usepackage{amsmath}
\usepackage{amssymb}
\usepackage{bm}
\usepackage{tabularx}
\usepackage{multirow}

\usepackage{mathtools}
\usepackage{amsfonts} 
\usepackage{siunitx}
\usepackage{algorithm,algorithmic}
\usepackage{booktabs}
\usepackage{threeparttable}
\usepackage{mathrsfs,amsmath}
\usepackage{multirow}
\usepackage{makecell}
\usepackage{placeins}
\usepackage{fixltx2e}
\usepackage{color}
\usepackage[pagebackref=true,breaklinks=true,letterpaper=true,colorlinks,bookmarks=false]{hyperref}

 \cvprfinalcopy % *** Uncomment this line for the final submission

\DeclareSIUnit\Molar{\textsc{m}} %captial M for molar in SI{}{}
\DeclareSIUnit{\pH}{pH}
\DeclareSIUnit{\pixel}{px}
\DeclareMathOperator*{\argminA}{arg\,min} % Jan Hlavacek
\DeclareMathOperator*{\argmaxA}{arg\,max} % Jan Hlavacek
\ifcvprfinal\fi
\begin{document}

%%%%%%%%% TITLE
\title{Multimodal Densenet}

\author{Faisal Mahmood$^1$, Ziyun Yang$^{1,2}$, Thomas Ashley$^1$ and Nicholas J. Durr$^1$\\
\\
$^1$Department of Biomedical Engineering, Johns Hopkins University, Baltimore, MD\\
$^2$Beijing Institute of Technology, Beijing, China\\
{\tt\small \{faisalm,ndurr\}@jhu.edu}}

\maketitle
%\thispagestyle{empty}

%%%%%%%%% ABSTRACT
\begin{abstract}
Humans make accurate decisions by interpreting complex data from multiple sources. Medical diagnostics, in particular, often hinge on human interpretation of multi-modal information. In order for artificial intelligence to make progress in automated, objective, and accurate diagnosis and prognosis, methods to fuse information from multiple medical imaging modalities are required. However, combining information from multiple data sources has several challenges, as current deep learning architectures lack the ability to extract useful representations from multimodal information, and often simple concatenation is used to fuse such information. In this work, we propose Multimodal DenseNet, a novel architecture for fusing multimodal data. Instead of focusing on concatenation or early and late fusion, our proposed architectures fuses information over several layers and gives the model flexibility in how it combines information from multiple sources. We apply this architecture to the challenge of polyp characterization and landmark identification in endoscopy. Features from white light images are fused with features from narrow band imaging or depth maps. This study demonstrates that Multimodal DenseNet outperforms monomodal classification as well as other multimodal fusion techniques by a significant margin on two different datasets.
\end{abstract}

%%%%%%%%% BODY TEXT
\section{Introduction}

\begin{figure}
\centering
\includegraphics[width=8cm]{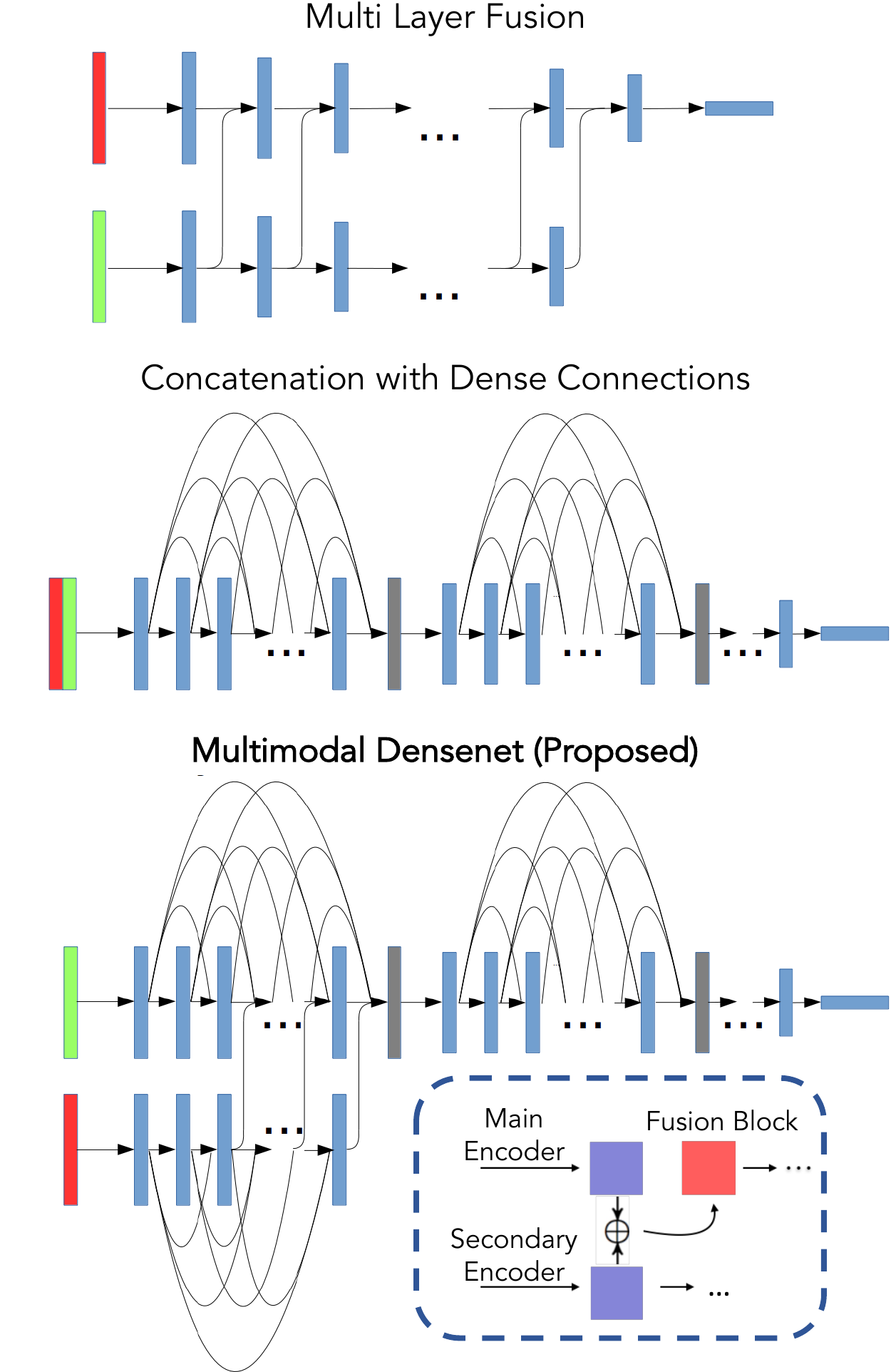}

\caption{A comparison of commonly used multimodal deep learning architectures with the proposed architecture. The state-of-the-art methods involve multi-layer fusion \cite{hazirbas2016fusenet,jiang2018rednet} or concatenation \cite{poria2015deep,sebastian2018bootstrapped}. We propose using a densenet style architecture and fusing the multimodal information in the last eight layers of the first dense block, hence harnessing the benefits of both multi-layer fusion and densely connected networks.}
\vspace{-5mm}
\end{figure}

Humans have evolved complex neurological machinery to make adept decisions based on information from many diverse sources of data. This process is critical for characterizing many medical conditions, where a diagnostic work-up may depend on information from patient history, physical examination, organ-level medical imaging, histological analysis, and laboratory studies. Even in a single medical imaging study, physicians commonly rely on multimodal contrast to determine an optimal diagnosis, for instance, considering both T1- and T2-weighted MRI images for identifying brain abnormalities. In screening for neoplasia in Barrett’s esophagus, a gastroenterologist can improve detection rates and decrease false positive findings by sequentially imaging the same tissue with high-resolution endoscopy, autofluorescence imaging, and narrow-band imaging \cite{curvers2008endoscopic}. Though the scenarios in which multimodal imaging are currently required and may stand the most to benefit from computer aided decision support, machine learning tools that effectively combine information from disparate sources remain immature and strategies for the optimal fusion of features from multimodal images is an active area of research.

\noindent\textbf{Contributions.} 
\begin{itemize}
%\vspace{-0.7em}
\item \textbf{DenseNet-based Multimodal Fusion:} We investigate a solution to fuse information from multiple modalities for more accurate scene classification in medical imaging applications. For this sake we propose a new architecture called Multimodal Densenet which is capable of fusing information from multiple modalities in a manner that fully harnesses dependencies between modalities and avoids common problems with vanishing gradients in multimodal networks. Our proposed architecture utilizes the strength of Densenet to tackle the vanishing gradiants problem and the first block of dense layers fuses the two sources of data (Fig. 1). 
\item \textbf{Unpaired Multimodal Data:} We further demonstrate that unpaired multimodal information from the same scene or organ can be aligned using adversarial training via a dual-GAN framework with cycle-consistency loss.
\item \textbf{Quantitative Study:} We comparatively analyze and quantitatively benchmark our proposed architecture with existing multimodal fusion architectures for scene classification such as modified version of FuseNet \cite{hazirbas2016fusenet} and RedNet \cite{jiang2018rednet} and demonstrate that our proposed architecture outperforms these methods for two different datasets. We present state-of-the-art results for lesion classification and anatomical landmark identification in colonoscopy, with data sources of white light endoscopy, narrow-band imaging, and depth. 
%%\vspace{-0.7em}
%\item To further validate the effectiveness of our approach we validate the depth estimation network with real endoscopy data collected from a porcine colon and its ground true depth acquired from a CT experiment.

\end{itemize}

\section{Related Work}

\noindent\textbf{Multimodal Deep Learning.} \\Multimodal machine learning continues to be an active research field with a variety of applications \cite{baltruvsaitis2018multimodal}. A major focus in early multimodal machine learning was on machine translation between visual data (e.g. images or video) and lingual data (\textit{e.g.} audio or text) \cite{ngiam2011multimodal,srivastava2012multimodal}. Later, Socher \textit{et. al.} demonstrated that multimodal feature modeling could be used to identify objects that a model had never seen before \cite{socher2013zero}.
As convolutional neural networks (CNNs) became popular for image classification \cite{krizhevsky2012imagenet}, they also started being used in multimodal tasks such as image captioning and text based image retrieval \cite{kiros2014multimodal}.

%Within this application, Ngiam et. al. and Srivastava et. al. were among the first groups to use deep learning to model multimodal features [Multimodal Deep Learning, Multimodal Learning with Deep Boltzmann Machines]. 

A common way to process temporally structured data is through the use of recurrent neural networks (RNNs) such as the long short term memory (LSTM) model \cite{hochreiter1997long}. For multimodal machine learning, Donahue \textit{et.al.} combined CNNs with LSTMs to caption videos \cite{donahue2015long}. A more recent trend in deep learning is using attention based models where the model can dynamically weigh different portions of a sequence of data \cite{ba2014multiple}. Attention based models have been used in multimodal applications as well \cite{yang2016stacked}.

Multimodal machine learning is also used for cross-modal data synthesis. These techniques are particularly prevalent in the medical field, as there are significant cost and privacy barriers in collecting medical data. For example, Huang \textit{et. al.} and Vemulapalli \textit{et. al.} were able to produce synthetic T2 weighted MRI images of the brain from T1 weighted MRI images and vice versa \cite{huang2018multimodal}.

\begin{figure}
\centering
\includegraphics[width=8cm]{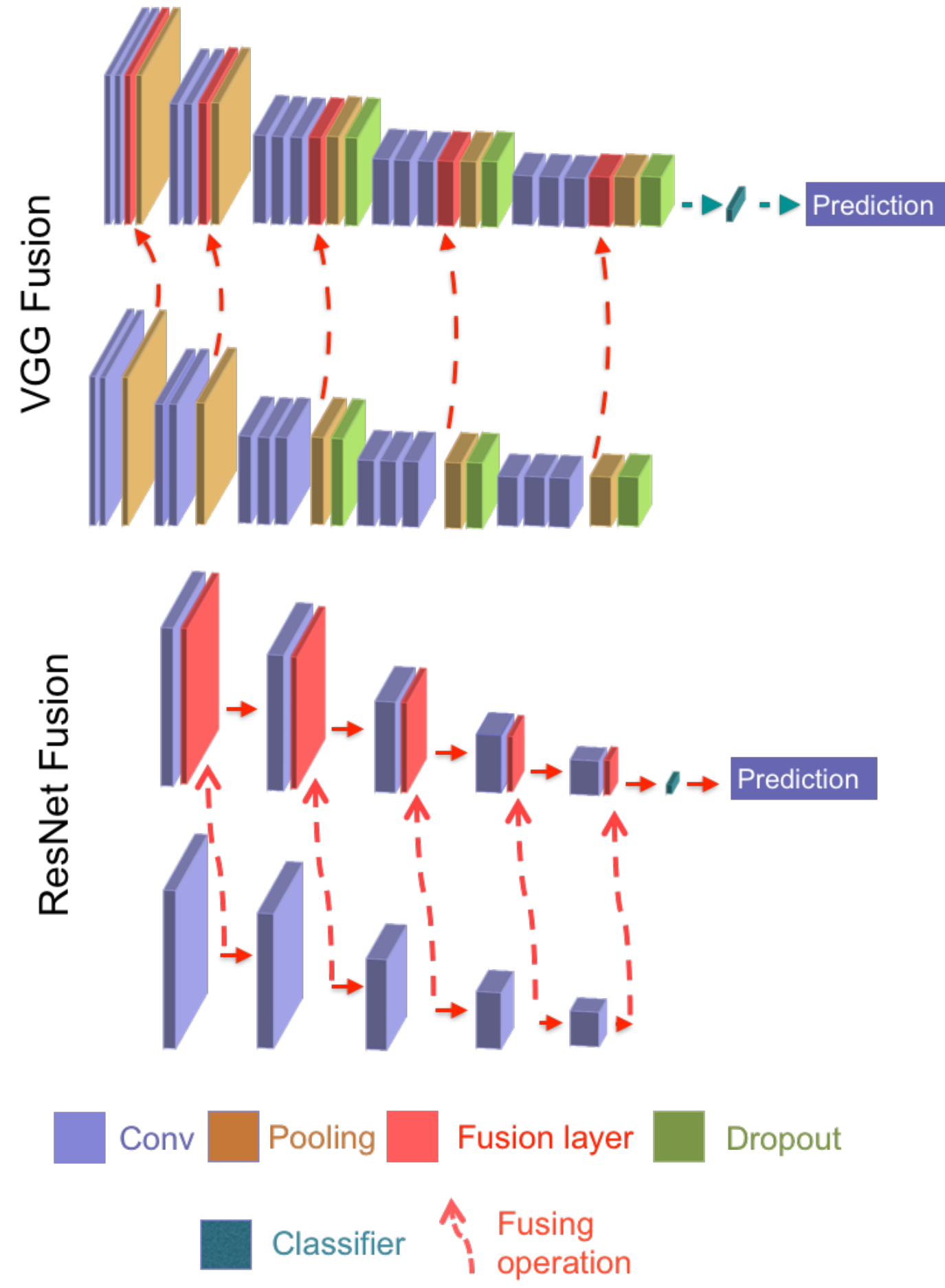}
\caption{Architectures representing fusion in VGG and ResNet, two branches of each network are used to extract features and those features are fused in a hierarchy as the network goes deeper. Fusion in feature space activates relevant neurons in both branches of the network. These two arrchitectures act as our baseline and are modified from FuseNet \cite{hazirbas2016fusenet} and RedNet \cite{jiang2018rednet}.}
\end{figure}

The advent of Generative Adversarial Networks (GANs) \cite{goodfellow2014generative} has expanded interest in synthetic data generation and domain adaptation \cite{mahmood2018unsupervised,mahmood2018deep} between modalities. Nie \textit{et. al.} use conditional GANs (cGANs) to produce synthetic CT images from MRI images \cite{nie2017medical}. Zhang \textit{et. al.} synthesize data in a similar task using both the real and synthetic data to improve heart segmentation \cite{zhang2018translating}. Shrivastava \textit{et. al.} \cite{shrivastava2017learning} use cGANS to make simulated images of the eye to appear more realistic, then use the realistic images to train a model to estimate eye gaze. Mahmood \textit{et. al.} reverse this method by using GANs to convert real medical images to simulation-like images in order to train a network to estimate depth in endoscopy \cite{mahmood2018unsupervised}.

\begin{figure*}
\centering
\includegraphics[width=\textwidth]{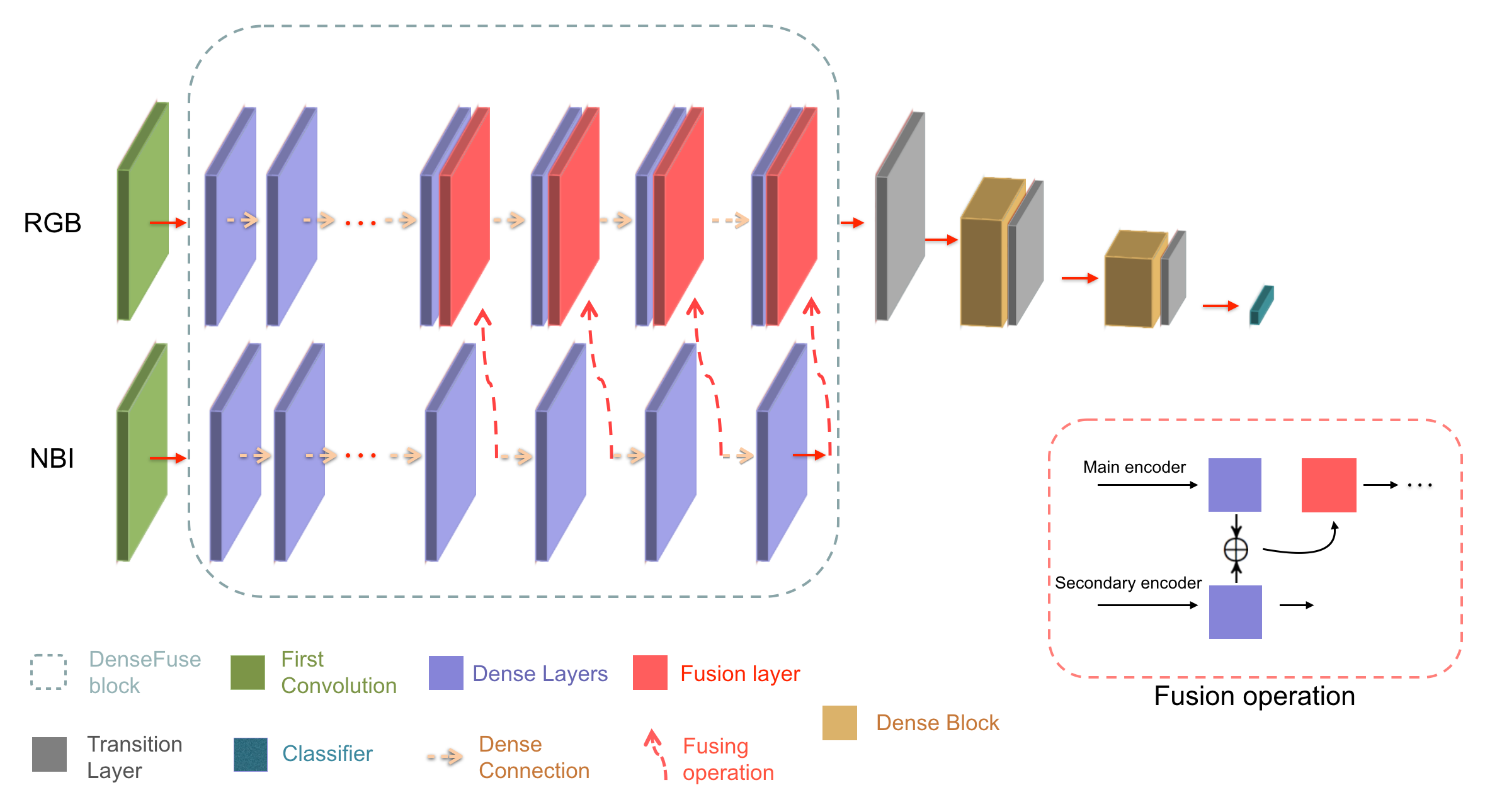}
\caption{The network is composed of three densely connected blocks of 16 dense layers each. The first block has a channel for each modality, and constantly fuses information from the bottom channel into the top through element-wise addition. Within the dense blocks, each dense layer is composed of two consecutive sets of Batch Normalization, ReLU, and convolutional layers. Between dense blocks, transition layers include average pooling to increase the receptive field of subsequent layers.}
\end{figure*}

\noindent\textbf{Data Fusion.} \\The use of multimodal data that is most relevant to our work is data fusion, in particular, fusing RGB and depth data for computer vision. One of the ways to classify data fusion methods is by early fusion, intermediate fusion, and late fusion. Generally, data is fused earlier when the modalities are correlated, and later when the modalities are less correlated \cite{ngiam2011multimodal}. Early, intermediate, and late fusion has all been used in various RGB-D applications. Eigen \textit{et. al.} used early fusion by concatenating RGB and depth features after a single layer in a network that performed semantic segmentation \cite{eigen2015predicting}. Long \textit{et. al.} used late fusion to perform RGB-D semantic segmentation \cite{long2015fully} and Eitel \textit{et. al.} used late fusion for RGBD object recognition \cite{eitel2015multimodal}. Roy \textit{et. al.} and Hazirbas \textit{et. al.} fuse RGBD data at multiple stages which could theoretically accommodate both low and high-level correlations between RGB and depth data \cite{roy2016multi,hazirbas2016fusenet}. Shahroudy \textit{et. al.} explicitly extract both correlated and uncorrelated features for action recognition in RGB-D video \cite{shahroudy2018deep}.
Use of depth information has not been limited to fusing data in a neural network. Qi \textit{et. al.} used depth data to construct a 3D graph that propagated non-local information and achieved state-of-the-art performance in semantic segmentation \cite{qi20173d}. Wang \textit{et. al.} modified the standard operations of convolution and pooling by weighing the contributions of neighboring pixels by their depth similarity \cite{wang2018depth}. This method also achieved competitive semantic segmentation results.

\noindent\textbf{Vanishing Gradients in Multimodal Networks.} \\The best performing CNNs for visual object recognition tasks contain many convolutional layers. One issue with deep models is the vanishing gradient problem, where the back-propagation update dwindles by the time it reaches the early layers. Song \textit{et. al.} demonstrated this problem in the context multimodal fusion and improved scene recognition by emphasizing the training of the bottom layers of a CNN \cite{song2017depth}. Another way to combat the vanishing gradient problem is by directly connecting lower layers with higher ones. Mao \textit{et. al.} implemented this idea by creating skip connections from the encoder to the decoder in their image restoration network \cite{mao2016image}. Similarly, densenet was designed to send the output of a layer directly to every layer after it \cite{huang2017densely}. This allows backpropagation to reach the bottom layers much earlier than in models without skip connections.

\section{Proposed Method: Multimodal Densenet}

\subsection{Theoratical Motivation}

Multi-layer fusion network such as \cite{hazirbas2016fusenet} are motivated by the fact that dual branch encoders with sequential feature fusion can pass a stronger signal which is representative of multiple modalities as compared to concatination which may loose information about a specific modality. In order to maintain a strong signal from one layer to another representing pertinent information from multiple sources, adding activations at corresponding layers instead of concatenation can be more fruitful. 

Heuristically, suppose the $k^{th}$ feature map in the $l^{th}$ layer is denoted by $\mathbf{h}_k^{(l)}$. Let the weights and biases for this feature and layer be denoted $W_K^{(l)} = [U_k^{(l)}\hspace{0.5em} V_K^{(l)}]^T$ and $b_k^{(l)} = [c_k^{(l)}\hspace{0.5em} d_k^{(l)}]^T$ respectively. Also, let $\mathbf{h} = \left[\mathbf{x}^T\hspace{0.5em}\mathbf{y}^T\right]^T$, where $\mathbf{x}$ and $\mathbf{y}$ are the learned features from the two modalities being fused. Assuming ReLU as the activation function,

\begin{align}
   \mathbf{h}_k^{(l+1)} &= \max(\mathbf{0}, W_k^{(l)}\mathbf{h} + y_k^{(l)}) \\
                         &= \max(\mathbf{0}, (U_k^{(l)}\mathbf{x}^{(l)} + V_k^{(l)}\mathbf{y}^{(l)} + (c_k^{(l)} + d_k^{(l)}))) \\
                         &\leq \max(\mathbf{0}, (U_k^{(l)}\mathbf{x}^{(l)} + c_k^{(l)})) + \max(\mathbf{0}, (V_k^{(l)}\mathbf{y}^{(l)} + d_k^{(l)})).
\end{align}

Based on the inequality shown in Eq. 3 we can see that the fusion of the activations in $\psi(x)$ and $\phi(y)$ produces a stronger signal than the activation on concatenated inputs. In the context of scene classification tasks the fusing operation results in the strongest signals being passed through the final later specifically at those places where the model finds useful information simultaneously in both modalities. In addition, the model preserves a signal, even though it may be a weaker one, when either modality contains useful information both for the learning of higher-level features and for the eventual classification of the image.

\subsection{Network Architecture}

The network architecture was conceived to utilize the strength of Densenet \cite{huang2017densely} for multimodal problems. The network is organized as a series of blocks, each composed of a series of 16 dense layers. Each dense layer is composed of the sequence of batch normalization, ReLU, bottleneck convolution, batch normalization, ReLU, and convolution. Height and width is preserved through a dense layer and the number of channels produced by the final convolution is known as the growth rate. Within the dense layer, the output from the bottleneck convolution is a fraction of the growth rate. The input to a dense layer is the concatenation of the outputs of all layers before it, within its block. This connection architecture combats the problem of vanishing gradients. Recent literature \cite{song2017depth} has emphasized the importance of training low layers in multimodal deep learning and dense connectivity should theoretically address this need.

The first block of dense layers fuses the two sources of data similar to \cite{hazirbas2016fusenet}. In this fuse block, there are two channels of dense layers, each processing an individual modality of data. During the first 8 dense layers, the RGB and depth channels proceed in parallel. Then, at each of the following 8 dense layers, the output from one secondary data channel is summed with the output of the primary data channel and is then used as input to the next layer in the primary channel. This is the key ingredient of the proposed multimodal architecture, such fusion combines the feature maps of the primary branch and the secondary branch. The rule of thumb of when to fuse multimodal data is to fuse early for highly correlated data and fuse late for uncorrelated data \cite{ramachandram2017deep}. For example, when fusing RGB and depth data they can be sometimes correlated, perhaps at an object’s edge, but sometimes uncorrelated, perhaps in the middle of a textured object, so its fusion should be an intermediate solution. Fusing two modalities over the course of several layers gives the model flexibility in how it combines information from multiple sources. 
The two blocks that come after the fuseblock consist of a single channel of 16 dense layers. Just between each block, there are transition layers that consist of batch normalization, ReLU, convolution, and average pooling which downsamples the data to half its size in both height and width. Finally, the data passes through a ReLU, average pooling, fully connected, and softmax layer to make a classification determination. Although, the architecture presented here is only for scene classification, it may be extended for other tasks such as semantic segmentation.

\subsection{Unpaired Multimodal Data}

\begin{figure*}
\centering
\includegraphics[width=\textwidth]{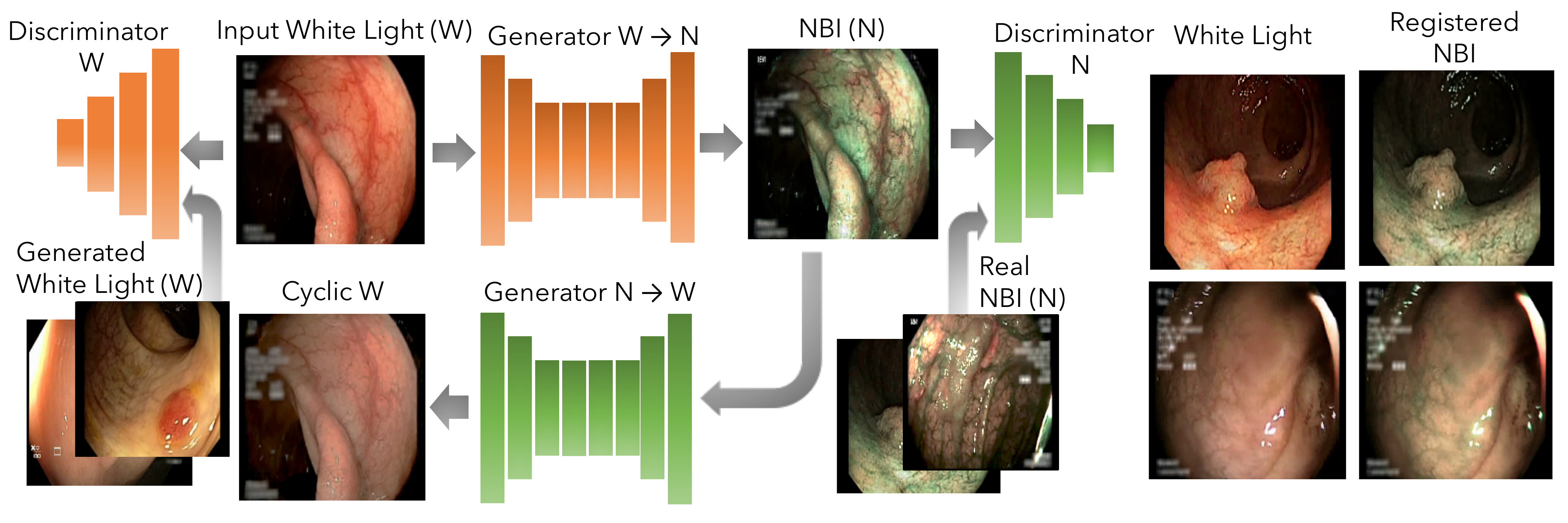}
\caption{Unpaired mapping between white light and NBI images trained on unpaired data from the same scene. The architecture involves a dual-GAN setup with cycle consistency loss \cite{zhu2017unpaired}, two generators learn a mapping $G: W\rightarrow N$ and $S:N\rightarrow W$ and two discriminators classify the data as real/fake pairs.}
\end{figure*}

Often the two modalities being used can be unpaired, even though they may have information from the same scene or organ they may not be registered. Since using supervised and mathematical registration techniques can be tedious we propose using generative adversarial networks for finding a mapping between the two modalities. We used cycle consistency loss \cite{zhu2017unpaired} with a conditional GAN to learn an unpaired mapping from white light to NBI images. Because of the consistency loss we also learn a reverse mapping form NBI images to white light images. Thus the reverse mapping is only used to train the forward mapping more effectively. Such an arrangement consists of four networks: $G$ (White light to NBI), $S$ (NBI to White light), $D_N$ (discriminator for $G$), and $D_W$ (discriminator for S). To train this framework for modality transfer with unpaired data, the overall objective consists of an adversarial loss term $\mathcal{L}_{\text{GAN}}$ and a cycle consistency loss term $\mathcal{L}_{\text{cyc}}$ to penalize unrealistic pairings between images. The adversarial loss is used to match the distribution of translated samples to that of the target distribution and can be expressed for both mapping functions. For the mapping $G: W \rightarrow N$ with discriminator $D_N$, we can express the objective as the binary cross entropy (BCE) loss of $D_N$ in classifying real or fake, in which $D_N$ and $G$ play a \textit{min-max} game to maximize and minimize this loss term respectively, the objective can be expressed as,

%$$ \text{min}_G \text{max}_{D_N} \mathcal{L}_{\text{GAN}}(G, D_{N}) = \mathbb{E}_{n\sim p_{\text{data}}(n)
%}[ \log D_{N}(n)]+\mathbb{E}_{m\sim p_{\text{data}}(m)}[\log(1-D_{N}(G(m)))] $$

%\argminA_G \argmaxA_{D_N} 
\begin{equation}
  \begin{aligned}
G^{*} \mathcal{L}_{\text{GAN}}(G, D_{N}) = \mathbb{E}_{n\sim p_{\text{data}}(n)}[ \log D_{N}(n)]\\+\mathbb{E}_{w\sim p_{\text{data}}(w)}[\log(1-D_{N}(G(w)))],
   \end{aligned}  
 \end{equation}

in which the generator $G$ aims to find a mapping between white light and NBI images, \textit{i.e.}, $G(m) \approx n$, while the discriminator $D_N$ aims to distinguish generated \textit{vs.} real NBI images. A similar objective can be expressed for $S: N \rightarrow M$,
%\argminA_S \argmaxA_{D_M} 
\begin{equation}
  \begin{aligned}
S^{*} \mathcal{L}_{\text{GAN}}(S, D_{W}) = \mathbb{E}_{w\sim p_{\text{data}}(w)}[ \log D_{W}(w)]\\+\mathbb{E}_{n\sim p_{\text{data}}(n)}[\log(1-D_{W}(S(n)))].
   \end{aligned}  
 \end{equation}

The cycle consistency loss is used to incentivize a one-to-one mapping between samples in $W$ and $N$, and that $G$ and $S$ act as inverse functions to each other. Specifically, the term ensures that the forward and back translations between the white light and NBI are lossless and cycle consistent, \textit{i.e.}, $S(G(w)) \approx w$ (forward cycle consistency) and $G(S(N)) \approx N$ (backwards cycle consistency). The forward cycle loss term helps with refining NBI to be more realistic and avoid occlusions.

%$$ \mathcal{L}_{cyc}(G,S) = \lambda_n \mathbb{E}_{n\sim p_{\text{data}}(n)
%}[||G(S(n)) - n||_1] + \lambda_m\mathbb{E}_{m\sim p_{\text{data}}(m)
%}[||S(G(m)) - m||_1] $$

\begin{equation}
  \begin{aligned}
\mathcal{L}_{cyc}(G,S) = \lambda_n \mathbb{E}_{n\sim p_{\text{data}}(n)}[||G(S(n)) - n||_1] \\+ \lambda_m\mathbb{E}_{m\sim p_{\text{data}}(w)}[||S(G(w)) - w||_1]
   \end{aligned}  
 \end{equation}

where $\lambda$ controls the importance of the forward and backward cycle constraints. For domain transfer between white light and NBI images, we relaxed the $\lambda_w$ term. The full objective for synthetic data generation can thus be written as,

\begin{equation}
  \begin{aligned}
G^*, S^* = \argminA_{G,S} \argmaxA_{D_N, D_W} \mathcal{L}_{\text{GAN}}(G, D_{N}) \\+ \mathcal{L}_{\text{GAN}}(S, D_{W}) + \mathcal{L}_{cyc}(G,S)
   \end{aligned}  
 \end{equation}

\begin{table*}
\centering
\begin{tabular}{|l|l|l|l|l|l|l|l|}
\hline
 \textbf{Method}& \textbf{Test Error $\downarrow$} & \textbf{Accuracy $\uparrow$} & \textbf{Recall $\uparrow$} & \textbf{Precision $\uparrow$} &  \textbf{Specificity $\uparrow$} &  \textbf{MCC $\uparrow$}& \textbf{F1 Score $\uparrow$} \\ \hline
VGG16                               &  0.265&  0.825&  0.705&  0.791&  0.855&  0.614&0.720\\ \hline
VGG16-Fuse                          &  0.160&  0.893&  0.789&  0.714&  0.937&  0.660&0.732\\ \hline
ResNet50                            &  0.157&  0.895&  0.802&  0.831&  0.903&  0.720&0.809\\ \hline
ResNet50-Fuse                       &  0.117&  0.922&  0.873&  0.874&  0.944&  0.818&0.874\\ \hline
DenseNet                            &  0.241&  0.840&  0.728&  0.849&  0.865&  0.664&0.740\\ \hline
\centering\textbf{Multimodal Densenet} & \textbf{0.069}&  \textbf{0.954}&  \textbf{0.922}&  \textbf{0.933}& \textbf{0.965} & \textbf{0.894} & \textbf{0.923} \\ \hline
\end{tabular}
\vspace{2mm}
 \caption{Comparative analysis of different RGB and Depth Fusion Methods for Polyp Classification Data (3 Class Scene Classification Problem). VGG16, ResNet-50 and DenseNet are monomodal RGB networks while VGG16-Fuse, ResNet50-Fuse and Multimodal densenet are multimodal RGB-D networks.}
\end{table*}
\begin{table*}
\centering
\begin{tabular}{|l|l|l|l|l|l|l|l|}
\hline
 \textbf{Method}& \textbf{Test Error $\downarrow$} & \textbf{Accuracy $\uparrow$} & \textbf{Recall $\uparrow$} & \textbf{Precision $\uparrow$} &  \textbf{Specificity $\uparrow$} &  \textbf{MCC $\uparrow$}& \textbf{F1 Score $\uparrow$} \\ \hline
VGG16                                  &  0.265&  0.825&  0.705&  0.791&  0.855&  0.614&0.720\\ \hline
VGG16-Fuse                             &  0.171&  0.834&  0.774&  0.823&  0.920&  0.658&0.722\\ \hline
ResNet50                               &  0.157&  0.895&  0.802&  0.831&  0.903&  0.720&0.809\\ \hline
ResNet50-Fuse                          &  0.128&  0.903&  0.852&  0.844&  0.918&  0.806&0.836\\ \hline
DenseNet                               &  0.241&  0.840&  0.728&  0.849&  0.865&  0.664&0.740\\ \hline
\centering\textbf{Multimodal Densenet} & \textbf{0.084}&  \textbf{0.926}&  \textbf{0.916}&  \textbf{0.927}& \textbf{0.944} & \textbf{0.877} & \textbf{0.908} \\ \hline
\end{tabular}
\vspace{2mm}
 \caption{Comparative analysis of different Fusion Methods for fusing white light and narrow band imaging data for Polyp Classification (3 Class Scene Classification Problem). VGG16, ResNet-50 and DenseNet are monomodal RGB white light networks while VGG16-Fuse, ResNet50-Fuse and Multimodal densenet are multimodal white light + narrow band imaging networks.}
\end{table*}
\begin{table*}
\centering
\begin{tabular}{|l|l|l|l|l|l|l|l|}
\hline
 \textbf{Method}& \textbf{Test Error $\downarrow$} & \textbf{Accuracy $\uparrow$} & \textbf{Recall $\uparrow$} & \textbf{Precision $\uparrow$} &  \textbf{Specificity $\uparrow$} &  \textbf{MCC $\uparrow$}& \textbf{F1 Score $\uparrow$} \\ \hline
VGG16                                  &  0.375            &  0.701           &  0.681       &  0.683       &  0.764    &  0.608  &0.648\\ \hline
VGG16-Fuse                             &  0.306            &  0.748           &  0.718       &  0.728       &  0.799    &  0.621  &0.716\\ \hline
ResNet50                               &  0.282            &  0.781           &  0.762       &  0.760       &  0.773    &  0.653  &0.697\\ \hline
ResNet50-Fuse                          &  0.224            &  0.811           &  0.825       &  0.814       &  0.814    &  0.668  &0.782\\ \hline
DenseNet                               &  0.211            &  0.796           &  0.772       &  0.821       &  0.728    &  0.733  &0.785\\ \hline
\centering\textbf{Multimodal Densenet} & \textbf{0.158}&  \textbf{0.898}&  \textbf{0.839}&  \textbf{0.895}& \textbf{0.913} & \textbf{0.837} & \textbf{0.891} \\ \hline
\end{tabular}
\vspace{2mm}
 \caption{Comparative analysis of different Fusion Methods for fusing RGB and depth data for Anatomical Landmark and Pathological Classification (6 Class Scene Classification Problem). VGG16, ResNet-50 and DenseNet are monomodal RGB networks while VGG16-Fuse, ResNet50-Fuse and Multimodal densenet are multimodal RGB-D networks.}
\vspace{-5mm}
\end{table*}
\vspace{-3mm}
\section{Experiments}
In this section, we evaluate the proposed multimodal network through extensive quantative experimentation for three different multimodal classification tasks:

\begin{itemize}
%\vspace{-0.7em}
\item Using white light RGB and corresponding depth for polyp classification in endoscopy data.
\item Using white light and narrow band imaging (NBI) images for polyp classification in endoscopy data.
\item Using white light RGB and corresponding depth images for anatomical landmark and pathology classification.
\end{itemize}
\vspace{-3mm}
\subsection{Datasets}

We use two publicly available endoscopy datasets with ground truth labels:\\\\
\noindent\textbf{ISIT-UMR Multimodal Polyp Classification Dataset.} \\We use the ISIT-UMR Multimodal classification dataset\footnote{\href{http://www.depeca.uah.es/colonoscopy_dataset/}{http:\/\/www.depeca.uah.es/colonoscopy\_dataset/}} that has been previously used for polyp classification tasks with classical machine learning methods \cite{mesejo2016computer}. The dataset consists of 76 polyps with both white light and NBI videos from the same polyp.  Each video orbits around a polyp capturing the texture and color information from different angles in both modalities. Each video has ground truth labels from histopathology and also has ground truth from multiple gastroenterologists. The dataset includes 3 classes of polyps: 15 serrated adenomas, 21 hyperplastic lesions and 40 adenoma. The length of each video is approximately 30 seconds and all videos are recorded at 26 frames/second. The white light and NBI videos were unregistered and were registered using the cycle-consistency adversarial paradigm described in the previous section. The depth for each frame was predicted using deep learning-based endoscopy depth estimation methods \cite{mahmood2018deepcr,mahmood2018topographical}. 53 videos were used for training and 23 videos were used for test. The entire dataset was downsampled to $256\times 256$ images for efficient processing. 

\noindent\textbf{Kvasir Landmark and Pathological Classification Dataset.} \\The Kvasir\footnote{\href{http://datasets.simula.no/kvasir/}{http://datasets.simula.no/kvasir/}} dataset consists of several anatomical landmarks and pathological findings, we use seven classes, three for landmark detection (Z-line, Pylorus, Cecum) and three for pathological findings (Esophagitis, Polyps and Ulcerative Colitis) and a class for Normal colon mucosa. The dataset contains $1,000$ unique images for each class, 7000 images were used in total. 70\% of the data was used for training 10\% for validation and 20\% for testing. Depths were estimated for each frame using deep learning-based methods \cite{mahmood2018deepcr,mahmood2018topographical}. 

\subsection{Implementation Details}

\noindent\textbf{General Implementation Details.} All networks were implemented using PyTorch 0.4 with Cuda 9.0. Google cloud with multiple Nvidia P100 GPUs was used to train all networks. All classification networks were trained using stochastic gradient decent (SGD) and images were down sampled $256\times 256$ for efficient processing and to the batch size was constant at 16 for 200 epochs. 

\noindent\textbf{DenseNet.} The network was trained for 200 epochs with an initial learning rate of 0.001 for the first 50 epochs and 0.0005 for another 50 epochs followed by 0.0002 for 80 epochs and eventually linearly decreased to zero over the remaining 20 epochs. The batch size was 16 and the weight decay parameter was set to 0.0001 and Nesterov momentum was 0.9. The dropout rate was set to 0.2.  

\noindent\textbf{Multimodal DenseNet.} The network was trained for 200 epochs with an initial learning rate of 0.0002 for the first 40 epochs and 0.00005 for another 40 epochs followed by 0.00001 for 90 epochs and eventually linearly decreased to zero over the remaining 30 epochs. The batch size was 16 and the weight decay parameter was set to 0.0001 and Nesterov momentum was 0.9. The dropout rate was set to 0.2. Given this model is memory critical naive implementations may result in memory inefficiencies and memory-efficient DeneNets \cite{pleiss2017memory} may be used for to reduce such a possibility.

\noindent\textbf{ResNet50 and ResNet50-Fuse.} The standard ResNet architecture was used and the architecture presented in Fig. 2 was used to train ResNet-50 fuse for all three scene classification tasks. Both networks ran for 200 epochs at an initial learning rate of 0.001 which was decreased to 0.0005 after 50 epochs and was further linearly decreased for the remaining epochs. Weight decay was set to 0.0001 and the Nesterov momentum was 0.9 for both networks.

\noindent\textbf{VGG16 and VGG16-Fuse.} The standard VGG16 architecture and the architecture presented in Fig. 2 was used to train VGG16 fuse for all three scene classification tasks. Both networks ran for 200 epochs at an initial learning rate of 0.002 which was decreased to 0.0005 after 60 epochs and was further linearly decreased for the remaining epochs. Weight decay was set to 0.0001 and the Nesterov momentum was 0.9 for both networks.

\noindent \textbf{Unpaired Multimodal Data Mapping Implementation.} The generator architectures contain two stride-2 convolutions, nine residual blocks and two functionally constrained convolutions with stride $\frac{1}{2}$. Reflection padding was used to minimize artifacts. The discriminator architecture was a simple classifier with three layers and the output was $70 \times 70$ with the aim to classify weather these overlapping patches were real or fake. As suggested in \cite{zhu2017unpaired} a patch level discriminator has fewer parameters and is more easily applicable to various image sizes.  We observed that larger size images needed more residual blocks for efficient convergence. The GAN training was stabilized to prevent mode collapse by using spectral normalization \cite{miyato2018spectral}. For all experiments $\lambda_n = 70$ and $\lambda_w = 10$. Adam solver \cite{kingma2014adam} was used to solve the optimization problem with a batch size of 1 which was experimentally determined. A total of $300$ epochs were used. The learning rate was set to $0.0002$ for the first $150$ epochs and linearly decayed to zero for the remaining $150$ epochs. In practice the objective function is divided by two when optimizing the discriminator \textit{i.e.}, the discriminator learns at a lower rate as compared to the generator. All networks were trained from scratch with no prior knowledge and weights were initialized from a Gaussian distribution with a mean and standard deviation of $0$ and $0.02$ respectively.

\begin{figure}
\centering
\includegraphics[width=9cm]{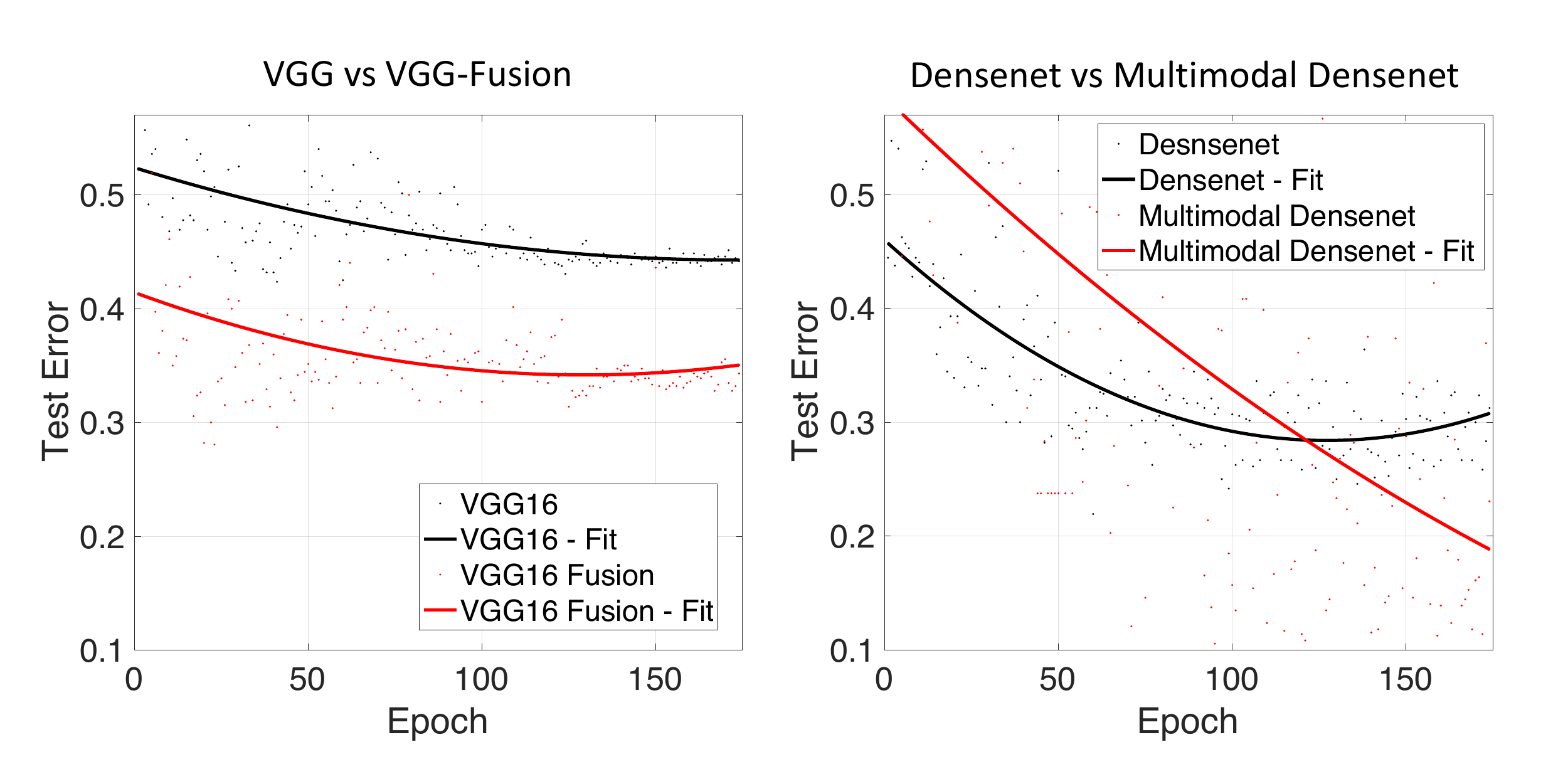}
\caption{A comparative analysis of VGG16 and VGG16 fusion (Fig. 2) vs Densenet and Multimodal Densenet. It can be seen that RGB-D classification works better than RGB classification in both networks. Although multimodal densenet needs more epochs to realize its full potential it outperforms VGG16 fusion.}
\end{figure}
\vspace{-4mm}
\begin{figure}
\vspace{-4mm}
\centering
\includegraphics[width=8cm]{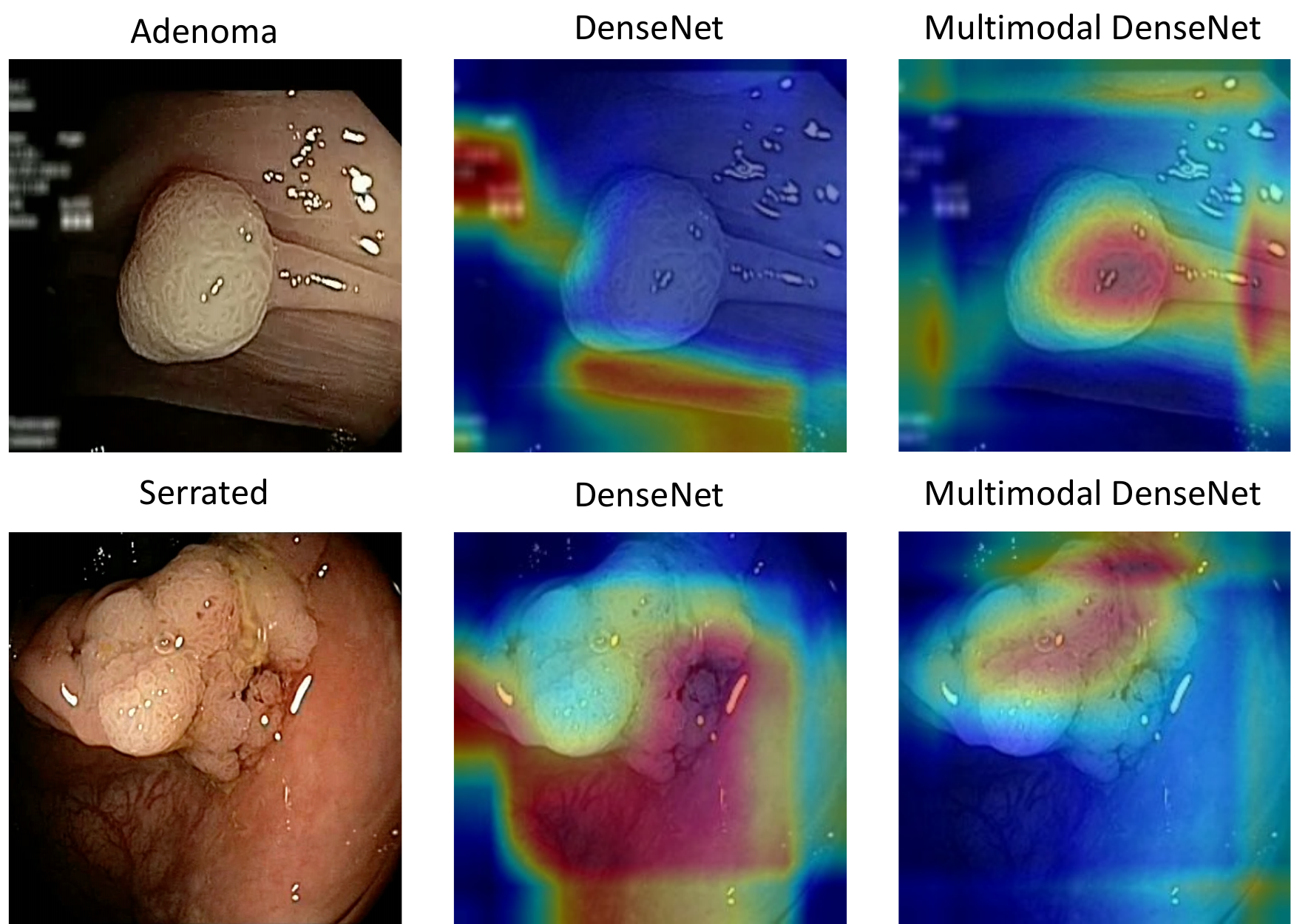}
\caption{Gradient Class Activation Maps (Grad-CAM) for Densenet (RGB) and Multimodal Densenet (RGB-D) showing regions used for classification when trained with the same amount of data. It can clearly be seen that Multimodal Densenet focuses on the polyps.}
\vspace{-4mm}
\end{figure}

\section{Results}
\vspace{-2mm}

\subsection{Evaluation Metrics}

We used seven different evaluation metrics to evaluate the three different classification problems. These include overall test error, accuracy, precision (ratio of samples that are correctly identified as positive), recall or sensitivity (ratio of samples that are correctly identified as positive among all existing positive samples), specificity (ratio of negatives that are correctly identified), Matthews correlation coefficient (MCC) \cite{matthews1985homeostasis}, and the F1 score (harmonic mean of the precision and recall). MCC was first proposed for binary problems but has since been extended for multi-class problems and is often referred to as the single most informative metric for classification problems \cite{chicco2017ten}.  
\vspace{-3mm}
\subsection{3-Class Polyp Characterization using RGB White Light images and Depth}
\vspace{-3mm}
During gastrointestinal screening physicians predict the depth of a polyp and subjectively assess the need for biopsy. Here we fuse information from predicted and white light RGB images and demonstrate that fusion depth can improve objective polyp classification. Table 1 shows the results for the three class classification problem with and without depth fusion using three different fusion networks. This comparative analysis is designed to compare standard multi-layer fusion networks. We do not compare by simply concatinating depth and RGB information because it has previously been shown that multi-layer fusion always out performs concatenation \cite{hazirbas2016fusenet,jiang2018rednet}. 

The $F_1$ score for RGB-D Multimodal Densenet is by 24.72\% greater than that for DenseNet. The percentage change for VGG16-Fuse as compare to VGG16 is 4.28\% and that for ResNet-50-Fuse as compared to ResNet-50 is 8.03\%. This clearly indicates that our proposed Multimodal Densenet is capable to extracting information from both RGB information and depth simultaneously as compared to other networks. 

A comparative analysis of the test error with respect to epochs has been shown in Fig. 5. The comparison is between VGG16 fusion and multimodal densenet. As demonstrated our proposed multimodal densenet outperforms VGG16, VGG16-Fusion and Densenets. Although, it should be noted that multimodal densenets do require more training as compared to monomodal and other relatively shallow multi-layer fusion counterparts.

\noindent\textbf{Gradient Class Activation Maps (Grad-CAM):} In order to further demonstrate the superiority of multimodal densenet over its monomodal counterpart and to determine the accuracy of our training we visualized the features being used to make classification determinations by our networks. This was done by visualizing gradient class activation maps for each class \cite{selvaraju2017grad}. It can be seen that given the small amount of data we used densenet often makes determinations based on regions around the polyp rather than directly on the surface of the polyp. However, when depth was fused using and the proposed multimodal architecture was use the classification label was determined from the surface of the polyp possibly using color and texture of the lumen. Indeed, in most cases densenet is capable of making the right determinations, this analysis only shows cases where RGB densenets fail and RGB-D multimodal densenet is capable of classifying the polyp accurately. 

\subsection{3-Class Polyp Characterization using White Light and NBI}

We further extend our quantitative analysis by exploring the fusion of narrow band imaging with white light endoscopy images. During gastrointestinal screening gastroenterologists often switch between white light and narrow band endoscopy to characterize a polyp and determine the need for a biopsy. Using the dataset described earlier we had videos from both modalities which were unpaired and unregistered. We registered them using the adversarial setup with cycle consistency loss described in section 3.3 (Fig. 4). For comparative analysis registered images from white light and narrow band imaging were fused using multimodal densenet, VGG16-Fuse and ResNet-50-Fuse baselines (Fig. 2). The results are shown in Table 2. The $F_1$ score for white light + NBI multimodal Densenet is by 8.03\% greater than that for DenseNet. The percentage change for VGG16-Fuse as compare to VGG16 is 3.33\% and that for ResNet-50-Fuse as compared to ResNet-50 is 1.66\%. This clearly indicates that our proposed multimodal densenet architecture is capable to extracting information from both white light and NBI simultaneously as compared to other fusion networks.

\subsection{7-Class Anatomical Landmarks and Pathological Classification}

Capsule endoscopy is an efficient method for upper gastrointestinal screening. However, the images acquired form this method are screened after the procedure is complete and the physician often spends several minutes watching a video in high speed, during this process it is possible to miss polyps and anatomical landmarks. Here, we should that it is possible to objectively classify anatomical landmarks and pathological findings using a 7-class classification problem. We further show that it is possible to fuse depth with RGB images to get improved results using the proposed multimodal densenet architecture (Table 3). 
\vspace{-2mm}
\noindent\section{Conclusions}
\vspace{-2mm}
In this paper, we present a new architecture to fuse information from multiple medical imaging modalities. We propose using a densenet style architecture and fusing the multimodal information in the last eight layers of the first dense block, hence harnessing the benefits of both multi-layer fusion and densely connected networks. We further demonstrate that unpaired information from the same scene or organ can be used for such fusion if the domains are aligned using adversarial training and cycle consistency loss. We demonstrate that our proposed architecture out performs existing fusion strategies and architectures for three multimodal scene classification tasks: (1) RGB-D polyp classification in endoscopy (2) White light-NBI polyp classification in endoscopy (3) RGB-D anatomical landmark and pathological classification in endoscopy.

{\small
\bibliographystyle{ieee}

\begin{thebibliography}{10}\itemsep=-1pt

\bibitem{ba2014multiple}
J.~Ba, V.~Mnih, and K.~Kavukcuoglu.
\newblock Multiple object recognition with visual attention.
\newblock {\em arXiv preprint arXiv:1412.7755}, 2014.

\bibitem{baltruvsaitis2018multimodal}
T.~Baltru{\v{s}}aitis, C.~Ahuja, and L.-P. Morency.
\newblock Multimodal machine learning: A survey and taxonomy.
\newblock {\em IEEE Transactions on Pattern Analysis and Machine Intelligence},
  2018.

\bibitem{chicco2017ten}
D.~Chicco.
\newblock Ten quick tips for machine learning in computational biology.
\newblock {\em BioData mining}, 10(1):35, 2017.

\bibitem{curvers2008endoscopic}
W.~L. Curvers, R.~Singh, L.~W.-K. Song, H.~C. Wolfsen, K.~Ragunath, K.~Wang,
  M.~B. Wallace, P.~Fockens, and J.~Bergman.
\newblock Endoscopic tri-modal imaging for detection of early neoplasia in
  barrett’s oesophagus: a multi-centre feasibility study using
  high-resolution endoscopy, autofluorescence imaging and narrow band imaging
  incorporated in one endoscopy system.
\newblock {\em Gut}, 57(2):167--172, 2008.

\bibitem{donahue2015long}
J.~Donahue, L.~Anne~Hendricks, S.~Guadarrama, M.~Rohrbach, S.~Venugopalan,
  K.~Saenko, and T.~Darrell.
\newblock Long-term recurrent convolutional networks for visual recognition and
  description.
\newblock In {\em Proceedings of the IEEE conference on computer vision and
  pattern recognition}, pages 2625--2634, 2015.

\bibitem{eigen2015predicting}
D.~Eigen and R.~Fergus.
\newblock Predicting depth, surface normals and semantic labels with a common
  multi-scale convolutional architecture.
\newblock In {\em Proceedings of the IEEE International Conference on Computer
  Vision}, pages 2650--2658, 2015.

\bibitem{eitel2015multimodal}
A.~Eitel, J.~T. Springenberg, L.~Spinello, M.~Riedmiller, and W.~Burgard.
\newblock Multimodal deep learning for robust rgb-d object recognition.
\newblock In {\em Intelligent Robots and Systems (IROS), 2015 IEEE/RSJ
  International Conference on}, pages 681--687. IEEE, 2015.

\bibitem{goodfellow2014generative}
I.~Goodfellow, J.~Pouget-Abadie, M.~Mirza, B.~Xu, D.~Warde-Farley, S.~Ozair,
  A.~Courville, and Y.~Bengio.
\newblock Generative adversarial nets.
\newblock In {\em Advances in neural information processing systems}, pages
  2672--2680, 2014.

\bibitem{hazirbas2016fusenet}
C.~Hazirbas, L.~Ma, C.~Domokos, and D.~Cremers.
\newblock Fusenet: Incorporating depth into semantic segmentation via
  fusion-based cnn architecture.
\newblock In {\em Asian Conference on Computer Vision}, pages 213--228.
  Springer, 2016.

\bibitem{hochreiter1997long}
S.~Hochreiter and J.~Schmidhuber.
\newblock Long short-term memory.
\newblock {\em Neural computation}, 9(8):1735--1780, 1997.

\bibitem{huang2017densely}
G.~Huang, Z.~Liu, L.~Van Der~Maaten, and K.~Q. Weinberger.
\newblock Densely connected convolutional networks.
\newblock In {\em CVPR}, volume~1, page~3, 2017.

\bibitem{huang2018multimodal}
X.~Huang, M.-Y. Liu, S.~Belongie, and J.~Kautz.
\newblock Multimodal unsupervised image-to-image translation.
\newblock {\em arXiv preprint arXiv:1804.04732}, 2018.

\bibitem{jiang2018rednet}
J.~Jiang, L.~Zheng, F.~Luo, and Z.~Zhang.
\newblock Rednet: Residual encoder-decoder network for indoor rgb-d semantic
  segmentation.
\newblock {\em arXiv preprint arXiv:1806.01054}, 2018.

\bibitem{kingma2014adam}
D.~P. Kingma and J.~Ba.
\newblock Adam: A method for stochastic optimization.
\newblock {\em arXiv preprint arXiv:1412.6980}, 2014.

\bibitem{kiros2014multimodal}
R.~Kiros, R.~Salakhutdinov, and R.~Zemel.
\newblock Multimodal neural language models.
\newblock In {\em International Conference on Machine Learning}, pages
  595--603, 2014.

\bibitem{krizhevsky2012imagenet}
A.~Krizhevsky, I.~Sutskever, and G.~E. Hinton.
\newblock Imagenet classification with deep convolutional neural networks.
\newblock In {\em Advances in neural information processing systems}, pages
  1097--1105, 2012.

\bibitem{long2015fully}
J.~Long, E.~Shelhamer, and T.~Darrell.
\newblock Fully convolutional networks for semantic segmentation.
\newblock In {\em Proceedings of the IEEE conference on computer vision and
  pattern recognition}, pages 3431--3440, 2015.

\bibitem{mahmood2018unsupervised}
F.~Mahmood, R.~Chen, and N.~J. Durr.
\newblock Unsupervised reverse domain adaptation for synthetic medical images
  via adversarial training.
\newblock {\em IEEE Transactions on Medical Imaging}, 2018.

\bibitem{mahmood2018deepcr}
F.~Mahmood, R.~Chen, S.~Sudarsky, D.~Yu, and N.~J. Durr.
\newblock Deep learning with cinematic rendering-fine-tuning deep neural
  networks using photorealistic medical images.
\newblock {\em Physics in Medicine and Biology}, 2018.

\bibitem{mahmood2018deep}
F.~Mahmood and N.~J. Durr.
\newblock Deep learning and conditional random fields-based depth estimation
  and topographical reconstruction from conventional endoscopy.
\newblock {\em Medical Image Analysis}, 2018.

\bibitem{mahmood2018topographical}
F.~Mahmood and N.~J. Durr.
\newblock Topographical reconstructions from monocular optical colonoscopy
  images via deep learning.
\newblock In {\em Biomedical Imaging (ISBI 2018), 2018 IEEE 15th International
  Symposium on}, pages 216--219. IEEE, 2018.

\bibitem{mao2016image}
X.-J. Mao, C.~Shen, and Y.-B. Yang.
\newblock Image restoration using convolutional auto-encoders with symmetric
  skip connections.
\newblock {\em arXiv preprint arXiv:1606.08921}, 2016.

\bibitem{matthews1985homeostasis}
D.~Matthews, J.~Hosker, A.~Rudenski, B.~Naylor, D.~Treacher, and R.~Turner.
\newblock Homeostasis model assessment: insulin resistance and $\beta$-cell
  function from fasting plasma glucose and insulin concentrations in man.
\newblock {\em Diabetologia}, 28(7):412--419, 1985.

\bibitem{mesejo2016computer}
P.~Mesejo, D.~Pizarro, A.~Abergel, O.~Rouquette, S.~Beorchia, L.~Poincloux, and
  A.~Bartoli.
\newblock Computer-aided classification of gastrointestinal lesions in regular
  colonoscopy.
\newblock {\em IEEE transactions on medical imaging}, 35(9):2051--2063, 2016.

\bibitem{miyato2018spectral}
T.~Miyato, T.~Kataoka, M.~Koyama, and Y.~Yoshida.
\newblock Spectral normalization for generative adversarial networks.
\newblock {\em arXiv preprint arXiv:1802.05957}, 2018.

\bibitem{ngiam2011multimodal}
J.~Ngiam, A.~Khosla, M.~Kim, J.~Nam, H.~Lee, and A.~Y. Ng.
\newblock Multimodal deep learning.
\newblock In {\em Proceedings of the 28th international conference on machine
  learning (ICML-11)}, pages 689--696, 2011.

\bibitem{nie2017medical}
D.~Nie, R.~Trullo, J.~Lian, C.~Petitjean, S.~Ruan, Q.~Wang, and D.~Shen.
\newblock Medical image synthesis with context-aware generative adversarial
  networks.
\newblock In {\em International Conference on Medical Image Computing and
  Computer-Assisted Intervention}, pages 417--425. Springer, 2017.

\bibitem{pleiss2017memory}
G.~Pleiss, D.~Chen, G.~Huang, T.~Li, L.~van~der Maaten, and K.~Q. Weinberger.
\newblock Memory-efficient implementation of densenets.
\newblock {\em arXiv preprint arXiv:1707.06990}, 2017.

\bibitem{poria2015deep}
S.~Poria, E.~Cambria, and A.~Gelbukh.
\newblock Deep convolutional neural network textual features and multiple
  kernel learning for utterance-level multimodal sentiment analysis.
\newblock In {\em Proceedings of the 2015 conference on empirical methods in
  natural language processing}, pages 2539--2544, 2015.

\bibitem{qi20173d}
X.~Qi, R.~Liao, J.~Jia, S.~Fidler, and R.~Urtasun.
\newblock 3d graph neural networks for rgbd semantic segmentation.
\newblock In {\em Proceedings of the IEEE Conference on Computer Vision and
  Pattern Recognition}, pages 5199--5208, 2017.

\bibitem{ramachandram2017deep}
D.~Ramachandram and G.~W. Taylor.
\newblock Deep multimodal learning: A survey on recent advances and trends.
\newblock {\em IEEE Signal Processing Magazine}, 34(6):96--108, 2017.

\bibitem{roy2016multi}
A.~Roy and S.~Todorovic.
\newblock A multi-scale cnn for affordance segmentation in rgb images.
\newblock In {\em European Conference on Computer Vision}, pages 186--201.
  Springer, 2016.

\bibitem{sebastian2018bootstrapped}
C.~Sebastian, B.~Boom, T.~van Lankveld, E.~Bondarev, and P.~H. De~With.
\newblock Bootstrapped cnns for building segmentation on rgb-d aerial imagery.
\newblock {\em arXiv preprint arXiv:1810.03570}, 2018.

\bibitem{selvaraju2017grad}
R.~R. Selvaraju, M.~Cogswell, A.~Das, R.~Vedantam, D.~Parikh, D.~Batra, et~al.
\newblock Grad-cam: Visual explanations from deep networks via gradient-based
  localization.
\newblock In {\em ICCV}, pages 618--626, 2017.

\bibitem{shahroudy2018deep}
A.~Shahroudy, T.-T. Ng, Y.~Gong, and G.~Wang.
\newblock Deep multimodal feature analysis for action recognition in rgb+ d
  videos.
\newblock {\em IEEE Transactions on Pattern Analysis and Machine Intelligence},
  40(5):1045--1058, 2018.

\bibitem{shrivastava2017learning}
A.~Shrivastava, T.~Pfister, O.~Tuzel, J.~Susskind, W.~Wang, and R.~Webb.
\newblock Learning from simulated and unsupervised images through adversarial
  training.
\newblock In {\em CVPR}, volume~2, page~5, 2017.

\bibitem{socher2013zero}
R.~Socher, M.~Ganjoo, C.~D. Manning, and A.~Ng.
\newblock Zero-shot learning through cross-modal transfer.
\newblock In {\em Advances in neural information processing systems}, pages
  935--943, 2013.

\bibitem{song2017depth}
X.~Song, L.~Herranz, and S.~Jiang.
\newblock Depth cnns for rgb-d scene recognition: Learning from scratch better
  than transferring from rgb-cnns.
\newblock In {\em AAAI}, pages 4271--4277, 2017.

\bibitem{srivastava2012multimodal}
N.~Srivastava and R.~R. Salakhutdinov.
\newblock Multimodal learning with deep boltzmann machines.
\newblock In {\em Advances in neural information processing systems}, pages
  2222--2230, 2012.

\bibitem{wang2018depth}
W.~Wang and U.~Neumann.
\newblock Depth-aware cnn for rgb-d segmentation.
\newblock {\em arXiv preprint arXiv:1803.06791}, 2018.

\bibitem{yang2016stacked}
Z.~Yang, X.~He, J.~Gao, L.~Deng, and A.~Smola.
\newblock Stacked attention networks for image question answering.
\newblock In {\em Proceedings of the IEEE Conference on Computer Vision and
  Pattern Recognition}, pages 21--29, 2016.

\bibitem{zhang2018translating}
Z.~Zhang, L.~Yang, and Y.~Zheng.
\newblock Translating and segmenting multimodal medical volumes with cycle-and
  shapeconsistency generative adversarial network.
\newblock In {\em Proceedings of the IEEE Conference on Computer Vision and
  Pattern Recognition}, pages 9242--9251, 2018.

\bibitem{zhu2017unpaired}
J.-Y. Zhu, T.~Park, P.~Isola, and A.~A. Efros.
\newblock Unpaired image-to-image translation using cycle-consistent
  adversarial networks.
\newblock {\em arXiv preprint}, 2017.

\end{thebibliography}

}

\end{document}